\documentclass[preprints,article,accept,pdftex,moreauthors]{Definitions/mdpi} 

\usepackage{ifthen}  

\makeatletter
\def\@arttype{} 
\makeatother

\usepackage{geometry}
\usepackage{newfloat}
\usepackage{caption}
\usepackage{seqsplit}
\newcommand{\leftcolUpdateCitation}
\firstpage{1} 
\pubvolume{1}
\issuenum{1}
\articlenumber{0}
\pubyear{2024}
\copyrightyear{2024}
\datereceived{ } 
\daterevised{ } 
\dateaccepted{ } 
\datepublished{ } 
\hreflink{https://doi.org/} 
\pdfoutput=1 

\Title{Foundational Models for Pathology and Endoscopy Images: 
Application for Gastric Inflammation}

\Author{Hamideh Kerdegari$^{1}$, Kyle Higgins$^{1,2}$, Dennis Veselkov$^{1}$, Ivan Laponogov$^{1}$, Inese Polaka$^{3}$, Miguel Coimbra$^{4,5}$, Junior Andrea Pescino$^{6}$, Mārcis Leja$^{3}$, Mário Dinis-Ribeiro$^{7}$, Tania Fleitas Kanonnikoff$^{8}$*, and Kirill Veselkov$^{1,9}$*}

\AuthorNames{Firstname Lastname, Firstname Lastname and Firstname Lastname}

\AuthorCitation{Lastname, F.; Lastname, F.; Lastname, F.}

\address{%
$^{1}$ \quad Division of Cancer, Department of Surgery and Cancer, Faculty of Medicine, Imperial College London, London, UK; h.kerdegari@imperial.ac.uk; k.higgins22@imperial.ac.uk; d.veselkov@imperial.ac.uk; i.laponogov@imperial.ac.uk; kirill.veselkov04@imperial.ac.uk \\
$^{2}$ \quad  Department of Neurobiology, Boston Children’s Hospital and Harvard Medical School, Boston, MA, USA; kyle.higgins@childrens.harvard.edu \\
$^{3}$ \quad  Faculty of Medicine, Institute of Clinical and Preventive Medicine, University of Latvia, Riga, Latvia; inese.polaka@lu.lv; marcis.leja@lu.lv \\
$^{4}$ \quad Instituto de Engenharia de Sistemas e Computadores, Tecnologia e Ciência, 3200-465 Porto, Portugal; mcoimbra@fc.up.pt \\
$^{5}$ \quad  Faculdade de Ciências, Universidade do Porto, 4169-007 Porto, Portugal; mcoimbra@fc.up.pt \\
$^{6}$ \quad  StratejAI, Avenue Louise 209, 1050 Brussels, Belgium; a.pescino@stratejai.com\\
$^{7}$ \quad  IRISE@CI-IPOP (Health Research Network), Portuguese Oncology Institute of Porto (IPO Porto), Portugal; mdinisribeiro@gmail.com \\
$^{8}$ \quad  Instituto Investigación Sanitaria INCLIVA, Medical Oncology Department, Hospital Clínico Universitario de Valencia, Valencia, Spain; tfleitas@incliva.es\\
$^{9}$ \quad  Department of Environmental Health Sciences, Yale University, New Haven, CT, USA; kirill.veselkov04@imperial.ac.uk}
\corres{Correspondence: kirill.veselkov04@imperial.ac.uk, tfleitas@incliva.es}

\abstract{The integration of artificial intelligence (AI) in medical diagnostics represents a significant advancement in managing upper gastrointestinal (GI) cancer, a major cause of global cancer mortality. Specifically for gastric cancer (GC), chronic inflammation causes changes in the mucosa such as atrophy, intestinal metaplasia (IM), dysplasia and ultimately cancer. Early detection through endoscopic regular surveillance is essential for better outcomes. Foundation models (FM), which are machine or deep learning models trained on diverse data and applicable to broad use cases, offer a promising solution to enhance the accuracy of endoscopy and its subsequent pathology image analysis. This review explores the recent advancements, applications, and challenges associated with FM in endoscopy and pathology imaging. We started by elucidating the core principles and architectures underlying these models, including their training methodologies and the pivotal role of large-scale data in developing their predictive capabilities. Moreover, this work discusses emerging trends and future research directions, emphasizing the integration of multimodal data, the development of more robust and equitable models, and the potential for real-time diagnostic support. This review aims to provide a roadmap for researchers and practitioners in navigating the complexities of incorporating FM into clinical practice for prevention/management of GC cases, thereby improving patient outcomes.}

\keyword{gastric cancer; endoscopy; pathology; foundation models} 

\begin{document}
\section{Introduction}\label{intro}
In this section, we first explain how AI can transform the detection and surveillance of upper gastrointestinal (GI) cancer, followed by a discussion on foundation models as a new era in medical imaging, specifically in endoscopy and pathology.

\subsection{AI in Upper GI Cancer: Transforming Detection and Surveillance}
\begin{figure}[H]\centering
\includegraphics[width=1\textwidth]{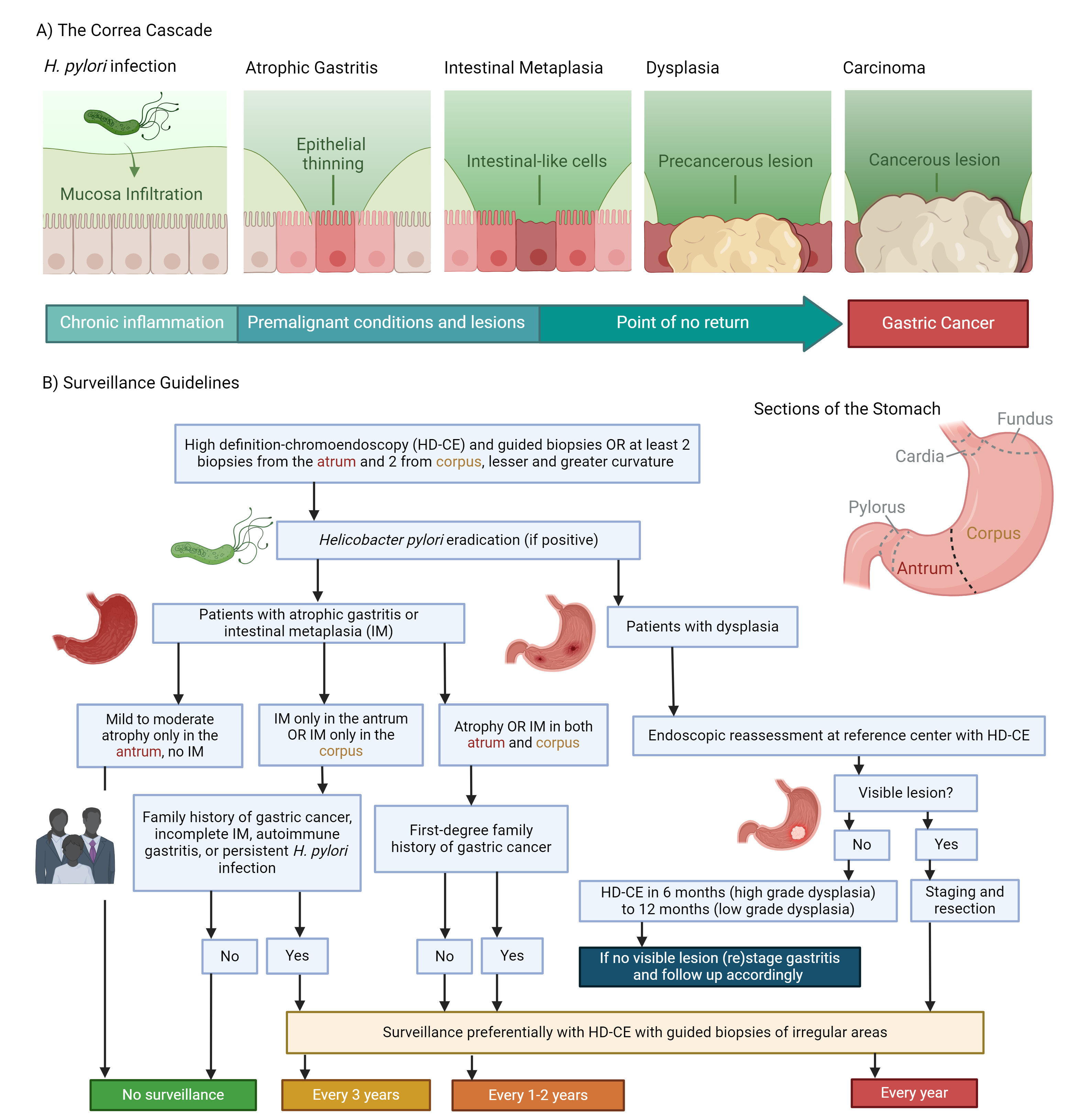}
\caption{A. The Correa’s cascade of intestinal type Gastric Carcinogenesis: a sequence of gastric changes  from chronic gastritis to atrophic gastritis, then to intestinal metaplasia and dysplasia, culminating in gastric cancer, highlighting a progressive, stepwise development toward malignancy. B. Surveillance guidelines overview \cite{pimentel2019management}: 1) Detection and Diagnosis: Endoscopy provides a direct view of the stomach lining, enabling the identification of areas that may exhibit precancerous alterations. During this examination, targeted biopsies are collected from visually abnormal or suspicious regions. 2) Pathological Analysis: These biopsies. are meticulously analyzed by pathologists to categorize the cellular composition of the tissue. This examination distinguishes between normal cells, atrophic gastritis, intestinal metaplasia, dysplasia, or the early stages of gastric cancer. The results are used for confirming the diagnosis and assessing the condition's severity. 3) Guiding Management: Insights derived from the endoscopic findings and pathological reports are integral to formulating a management strategy. Decisions regarding the frequency of surveillance, the need for further medical interventions, and evaluations of the risk for progression to gastric cancer are based on these combined observations and individual risk factors such as genetic predispositions.
\label{fig1}}
\end{figure}

Gastric cancer (GC) is one of the leading causes of cancer mortality globally. It’s pathogenesis is related with chronic inflammation which causes changes in the mucosa including atrophy, intestinal metaplasia (IM), and dysplasia \cite{yoon2015diagnosis} as shown in Figure~\ref{fig1}.A. Recognising these conditions early through regular and precise endoscopic surveillance as 
\noindent presented in Figure~\ref{fig1}.B is pivotal for enhancing early diagnosis and treatment outcomes \cite{pimentel2019management, matysiak2020recent}. The advent of artificial intelligence (AI) in medical imaging and diagnostics brings a promising solution to these challenges, especially in the realm of GI endoscopy. AI and FM offer a revolutionary approach to interpreting endoscopy and pathology images for risk stratification and determination of  the appropriate surveillance intervals for patients with upper GI precancerous conditions. These AI-driven models can potentially transform the management of patients at risk for upper GI cancers by providing precise, real-time analysis of endoscopic images, identifying premalignant lesions with high accuracy, and predicting the risk levels of patients based on the characteristics of detected lesions. This emphasis on AI's role in enhancing the identification and surveillance of high-risk patients marks a significant step forward. By leveraging AI for the interpretation of endoscopic and pathology images, healthcare providers can achieve more accurate risk stratification and timely intervention, ultimately aiming to increase the surveillance rate among high-risk patients. This not only addresses the current shortfall in surveillance adherence but also paves the way for a more proactive and prevention-oriented approach in managing the risk of upper GI cancers.

\subsection{Expanding horizons: Foundation Models in Endoscopy and Pathology Imaging}
The advent of FM has marked a pivotal shift in the landscape of medical imaging analysis, particularly in the domains of pathology and endoscopy. These models undergo training on large and varied datasets, often employing self-supervision methods on an extensive scale. After this initial training, they can be further refined—through processes like fine-tuning—to perform a broad spectrum of related downstream tasks, enhancing their applicability based on the original dataset. By leveraging vast amounts of data to learn rich representations, they have the potential to facilitate the diagnosis, for a personalized treatment approach. This review aims to explore the cutting-edge advancements in FM applied to pathology and endoscopy imaging, elucidating their impact, challenges, and the promising avenues they pave for future research.

\begin{figure}[h]
\centering
\includegraphics[width=\textwidth]{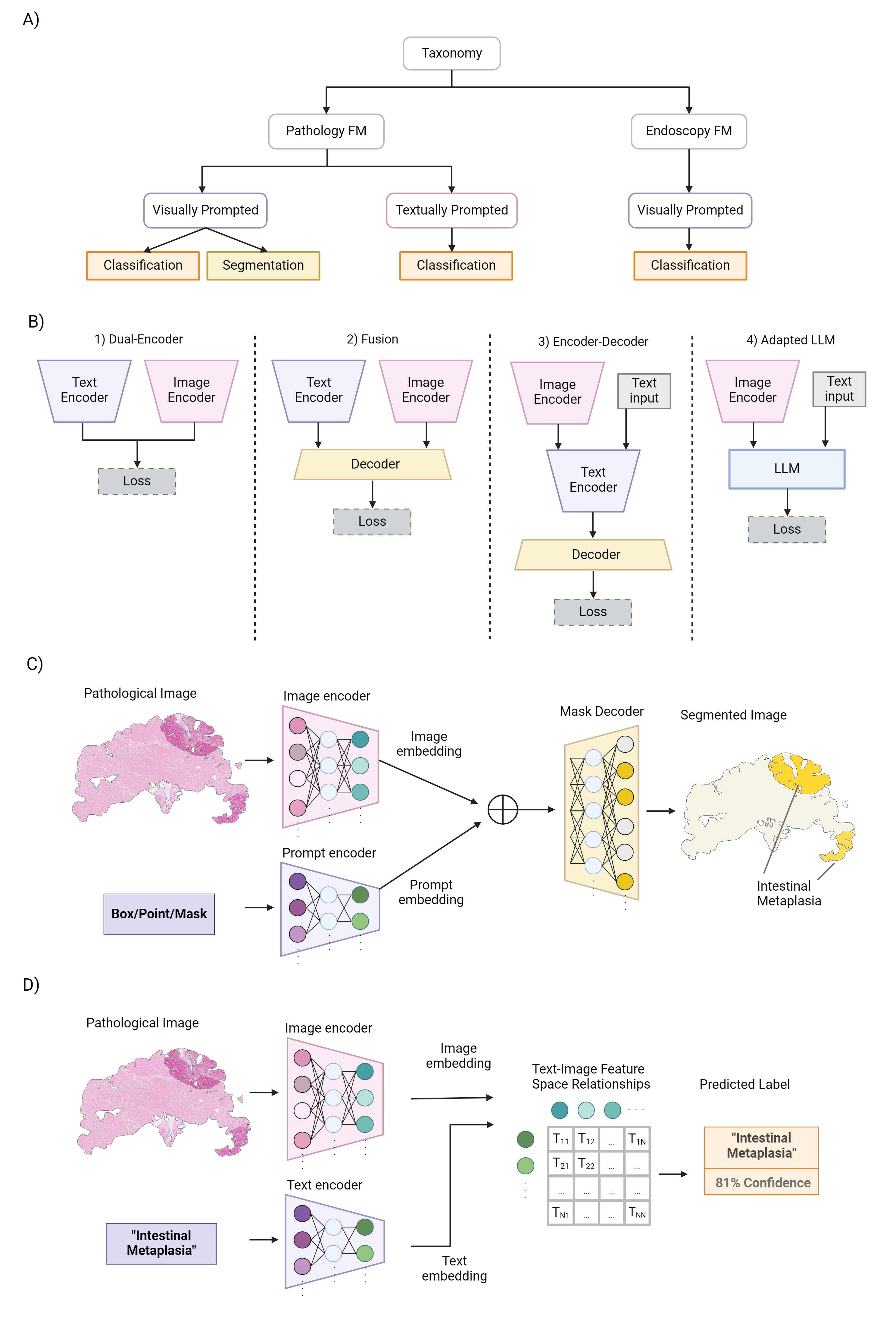}
\caption{ A) An overview of our taxonomy for pathology and endoscopy FM. They are categorized based on the prompt types (i.e visually or textually prompted models) and their utilization. B) Overview of four different common architecture styles used in vision language models: 1) Dual-Encoder designs use a parallel image and text encoder with aligned representations, 2) Fusion designs jointly process both image and text representations via a decoder, 3) Encoder-Decoder designs apply joint feature encoding and decoding sequentially, 4) Adapted large language model (LLM) designs input visual and text prompts to the LLMs to leverage their superior generalization ability. C) Overview of segment anything model (SAM) for pathology image segmentation. D) The process of training textually prompted models with paired image–text dataset via contrastive learning.
\label{fig22}}
\end{figure}

Pathology and endoscopy, the two critical fields in medical diagnostics, generate a wealth of image data that encapsulate intricate details vital for accurate disease diagnosis and management. Traditionally, the analysis of such images has been heavily reliant on the expertise of highly trained professionals. While indispensable, this approach is time-consuming, subject to variability, and scales linearly with the volume of data. FM emerge as a powerful solution to these challenges, offering a way to automate and enhance the analysis process through textually and visually prompted models. In the context of pathology, textually prompted models leverage textual data as prompts to guide the analysis of images. These textual prompts could be descriptions of histological features, diagnostic criteria, or other relevant annotations that guide the model's interpretation of the images. On the other hand, visually prompted models operate by utilizing visual cues, such as points, boxes, or masks, to guide the model's focus within an image. For a pathology image, a visually prompted model could be prompted to concentrate on specific areas of a tissue slide that are marked by a pathologist or identified by preliminary analysis. In the context of endoscopy, only visually prompted models are utilized since clinical routines for endoscopy videos do not involve text data. However, the integration of FM into pathology and endoscopy poses unique challenges. These include ensuring model interpretability, managing the privacy and security of sensitive medical data, and addressing the potential biases inherent in the training datasets. Moreover, the dynamic nature of medical knowledge and the continuous evolution of diseases necessitate that these models are adaptable and capable of learning from new data incorporating previously acquired knowledge. This paper provides an introductory overview, followed by an in-depth analysis of the principles underlying FM for medical imaging, with a focus on pathology and endoscopy (see Figure~\ref{fig22}.A for a taxonomy of vision-language FM). We classify the current state-of-the-art (SOTA) models based on their architectural designs, training objectives, and application areas. Furthermore, we discuss the recent works in the field, highlighting the innovative approaches that have been developed to address the specific needs of pathology and endoscopy image analysis. Finally, we outline the challenges faced by current models and propose several directions for future research, aiming to guide and inspire further advancements in this rapidly evolving field. Our major contributions include:
\begin{enumerate}
    \item We conduct a thorough review of FM applied in the field of pathology and endoscopy imaging, beginning with their architecture types, training objectives, and large-scale training. Then, they are classified  into visually and textually prompted models (based on prompting type), and then their subsequent application/utilization is discussed.
    \item We also discuss the challenges and unresolved aspects linked to FM in pathology and endoscopy imaging.

\end{enumerate}

\section{Inclusion and search criteria}
For this review, we have focused on identifying and evaluating studies related to the application of FM in pathology and endoscopy imaging. We included original research articles, review articles, and preprints that discuss their development, validation, and application. Only articles published in English were included to maintain consistency and accessibility of the content. We selected studies specifically addressing the use of FM in the detection, diagnosis, and management of GI conditions, including gastric inflammation and cancer. The databases searched included PubMed, Scopus, IEEE Xplore, Web of Science, Google Scholar, and specialized databases like Embase for medical and biomedical literature. Keywords and Medical Subject Headings (MeSH) terms used included "foundation models," "deep learning," "machine learning," "pathology," "endoscopy," "AI diagnostics," "pretrained models," and "transfer learning." Boolean operators were utilized to refine the search: AND was used to combine different concepts (e.g., "foundation models" AND "pathology"), OR to include synonyms or related terms (e.g., "endoscopy" OR "gastroscopy"), and NOT to exclude unwanted terms if necessary. Filters were applied for human studies, years of publication, and types of articles, excluding editorials and reviews if focusing on primary research. 

The search strategy involved a comprehensive review of electronic databases using the specified keywords. Initial search results were screened based on titles and abstracts to assess relevance, and full texts of potentially relevant articles were retrieved and reviewed. References from these articles were also examined to identify additional relevant studies. Comprehensive searches were conducted in the mentioned databases, and titles and abstracts were screened to identify studies meeting the inclusion criteria. Articles that did not meet these criteria were excluded at this stage. Full texts of the remaining articles were reviewed for relevance and inclusion, with any discrepancies in study selection resolved through discussion among the authors. Studies involving human subjects undergoing pathology analysis or endoscopy and those using FM on pathological data and endoscopic images or videos were included. Only studies employing FM, such as large-scale pretrained machine learning models that are then adapted or fine-tuned for specific tasks in pathology and endoscopy, were considered. By applying these inclusion and search criteria, we aimed to provide a comprehensive and focused review of the current state and future directions of FM in pathology and endoscopy imaging, specifically for the application in gastric inflammation and cancer.

\clearpage 

\section{Foundational Models in Computer Vision}\label{FMC}
\subsection{Architecture Type}

 Vision language models primarily utilize four distinct architectural frameworks, as depicted in Figure~\ref{fig22}.B. The initial framework, known as Dual-Encoder, employs parallel visual and textual encoders to produce aligned representations. The second framework, Fusion, integrates image and text representations through a fusion decoder, facilitating the learning of combined representations. The third framework, Encoder-Decoder, features a language model based on encoder-decoder mechanisms alongside a visual encoder, enabling sequential joint feature encoding and decoding. Finally, the Adapted LLM framework incorporates an LLM as its foundational element, with a visual encoder that transforms images into a format that LLM can understand, thus capitalizing on the LLM's enhanced generalization capabilities. Following this overview, we will explore the loss functions utilized to train these various architectural types.

\subsection{Training objectives}\label{TO}
\subsubsection{Contrastive Objectives}

Contrastive objectives train models to form distinct representations, effectively narrowing the gap between similar sample pairs and widening it between dissimilar ones in the feature space \cite{chen2020simple}. Image contrastive loss is designed to enhance the uniqueness of image features. It achieves this by aligning a target image more closely with its positive keys—essentially, versions of itself that have undergone data augmentation—and ensuring that it remains clearly differentiated from its negative keys, which are distinct, unrelated images, within the embedding space. Image-Text Contrastive (ITC) loss is a type of contrastive loss function that aims to create distinctive image-text pair representations. This is accomplished by bringing together the embeddings of matched images and texts and pushing apart those that do not match \cite{radford2021learning}. Given a batch of N examples, ITC loss aims to match correct image-text pairs among N $\times$ N possible configurations.
ITC loss maximizes cosine similarity between N correct pairs and minimizes it among \(N^2 - N\) incorrect pairs. Let $(x_i, t_i)$ be i-th image-text example and $(v_i, t_i)$ be its corresponding representations, then Image-Text Contrastive (ITC) loss is calculated as follows:
\begin{equation}
    \text{loss}_{ITC} = - \log \left[ \frac{\exp(\text{sim}(v_i, t_i)/\tau)}{\sum_{j=1}^{n} \exp(\text{sim}(v_i, t_j)/\tau)} \right]
\end{equation}
This loss is calculated by concentrating on the relationship between images and texts, taking into account the temperature parameter $\tau$. ITC loss was used by \cite{jia2021scaling, radford2021learning} to learn to predict correct image-text pairs. 

Image-Text Matching (ITM) loss \cite{li2021align} is another type of contrastive loss that aims to correctly predict whether a pair of images and text is positively or negatively matched. 
 To achieve this, a series of perceptron layers are introduced to estimate the likelihood of a pair being matched. Subsequently, the loss is computed using the cross-entropy loss function. Additionally, several other contrastive loss functions have been utilized for various applications. They include variants of ITC losses such as FILIP loss \cite{yao2021filip}, Text-to-Pixel Contrastive loss \cite{wang2022cris}, Region-Word Alignment \cite{li2022grounded}, Multi-label Image-Text Contrastive \cite{xu2022groupvit}, Unified Contrastive Learning \cite{yang2022unified}, Region-Word Contrastive loss \cite{zhang2022glipv2}, and image-based self-supervision loss like simple contrastive learning of representations (SimCLR) \cite{chen2020simple}.
 
\subsubsection{Generative Objectives}\label{GO}
Generative objectives focus on training networks to create images or textual contents, enabling them to learn semantic attributes through activities such as image generation \cite{bao2021beit} and language production \cite{liu2019roberta}. A prevalent generative loss function in computer vision is Masked Image Modeling (MIM) \cite{zhang2023toward}. This approach involves the acquisition of cross-patch correlations by employing masking and image reconstruction methods. In MIM, certain patches of an input image are randomly obscured, and the network's encoder is tasked with reconstructing these hidden patches using the visible sections as a reference. For a given batch of N images, the loss function is computed as follows:
\begin{equation}
LMIM = -\frac{1}{N} \sum_{i=1}^{N} \log f_{\theta}\left( x_i^{-I} | x_i^{\wedge I}\right)
\end{equation}

where $x_i^{-I}$ and $x_i^{\wedge I}$ represent the masked and unmasked patches within $x_i^{I}$, respectively. Similarly, Masked Language Modeling (MLM) \cite{zhang2023toward} is a widely adopted generative loss technique in Natural Language Processing (NLP). In MLM, a certain percentage of input text tokens are randomly masked, and the model is trained to predict these masked tokens based on the context provided by the unmasked tokens. The loss function used in MLM is akin to that of the MIM, focusing on the reconstruction of masked elements. However, the difference lies in the elements being reconstructed: in MLM, the focus is on masked and unmasked tokens, as opposed to the masked and unmasked patches in MIM. In similar fashion, various other generative loss techniques have been proposed. Examples include Masked Multimodal Modeling (MMM) loss \cite{singh2022flava}, Semi-Casual Language Modeling \cite{hao2022language}, Image-conditioned Masked Language Modeling (IMLM) loss \cite{zhang2023toward}, Image-grounded Text Generation (ITG) loss \cite{li2023blip2}, and Captioning with Parallel Prediction (CapPa) loss \cite{tschannen2023image}.

\subsection{Large-scale Training}\label{LST}
Large-scale data is central to the development of vision and language foundational models. The datasets used for pre-training these models are categorized into three main types: image-text datasets, partially synthetic datasets, and combination datasets. Among these, image-text datasets like WebImageText, which are used in CLIP \cite{radford2021learning}, demonstrate the significant impact of web-scale image-text data in training FM. Such data is typically extracted from web crawls and the final dataset emerges from a rigorous filtering process designed to eliminate noisy, irrelevant, or detrimental data points. Unlike image-text datasets, partially synthetic datasets are not readily available on the web and require significant human annotation effort. A cost-effective approach is to utilize a proficient teacher model to transform image-text datasets into mask-description datasets \cite{li2022grounded}. To address the challenge of curating and training on web-scale datasets, combination datasets have been employed \cite{chen2020uniter, tsimpoukelli2021multimodal, xu2022unifying}. These datasets amalgamate standard vision datasets, including those featuring image-text pairs such as captioning and visual question answering, and sometimes modify non-image-text datasets using template-based prompt engineering to transform labels into descriptions. 

Furthermore, large-scale training, coupled with effective fine-tuning and strategic prompting at the inference stage, has been an essential component of vision foundational models. Fine-tuning adjusts the model's parameters on a task-specific dataset, optimizing it for particular applications like image captioning or visual question answering. It enables leveraging the vast knowledge captured by pre-trained vision-language models, making them highly effective for a wide range of applications with relatively less data and computational resources than required for training from scratch. Prompt engineering, meanwhile, involves the strategic creation of input prompts to guide the model in generating accurate and relevant responses, leveraging its pre-trained capabilities. Both practices are vital for customizing vision language models to specific needs, enabling their effective application across various domains that require nuanced interpretations of visual and textual information.

\section{Pathology Foundation Models}\label{PFM}
\subsection{Visually Prompted Models}\label{VPM}
This section reviews visually prompted pathology foundational models that have been designed for segmentation and classification of pathology images.

\subsubsection{Pathology Image Segmentation}\label{PIS}
Semantic segmentation plays a crucial role in digital pathology, involving the division of images into distinct regions that represent different tissue structures, cell types, or subcellular components. The segment anything model (SAM) \cite{kirillov2023segment} emerges as the first promptable foundation model specifically designed for image segmentation tasks. Trained on the SA-1B dataset, SAM benefits from a vast amount of images and annotations, granting it outstanding zero-shot generalization capabilities. Utilizing a vision transformer-based image encoder, SAM extracts image features and computes image embeddings. Additionally, its prompt encoder embeds user prompts, enhancing interaction. The combined outputs from both encoders are then processed by a lightweight mask decoder, which generates segmentation results by integrating the image and prompt embedding with output tokens as illustrated in Figure~\ref{fig22}.C. While SAM has proven effective for segmenting natural images, its potential for navigating the complexities of medical image segmentation, particularly in pathology, invites further investigation. Pathology images present unique challenges, such as structural complexity, low contrast, and inter-observer variability. To address these, the research community has explored various extensions of SAM, aiming to unlock its capabilities for pathology image segmentation tasks.

For example, Deng et al. \cite{deng2023segment} evaluated SAM in the context of cell nuclei segmentation on whole slide imaging (WSI). Their evaluation encompassed various scenarios, including the application of SAM with a single positive point prompt, with 20 point prompts comprising an equal number of positive and negative points, and with comprehensive annotations (points or bounding boxes) for every individual instance. The findings indicated that SAM delivers exceptional performance in segmenting large, connected objects. However, it falls short in accurately segmenting densely packed instances, even when 20 prompts (clicks or bounding boxes) are used for each image. This shortfall could be attributed to the significantly higher resolution of WSI images relative to the resolution of images used to train SAM, coupled with the presence of tissue types of varying scales in digital pathology. Additionally, manually annotating all the boxes during inference remains time-consuming. To address this issue, Cui et al. \cite{cui2023allin} introduced a pipeline for label-efficient finetuning of SAM, with no requirement for annotation prompts during inference. Such a pipeline surpasses previous SOTA methods in nuclei segmentation and achieves competitive performance compared to using strong pixel-wise annotated data. Zhang et al. \cite{zhang2023sampath} introduced SAM-Path for semantic segmentation of pathology images. This approach extends SAM by incorporating trainable class prompts, augmented further with a pathology-specific encoder derived from a pathology FM. SAM-Path improves upon SAM's capability for performing semantic segmentation in digital pathology, eliminating the need for human-generated input prompts. The findings highlight SAM-Path's promising potential for semantic segmentation tasks in pathology. In another study, CellSAM \cite{israel2023foundation}, proposed as a foundational model for cell segmentation, generalizes across a wide range of cellular imaging data. This model enhances the capabilities of SAM by introducing a novel prompt engineering technique for mask generation. To facilitate this, an object detector named CellFinder was developed, which automatically detects cells and cues SAM to produce segmentations. They demonstrated that this approach allows a single model to achieve SOTA performance in segmenting images of mammalian cells (both in tissues and cell cultures), yeast, and bacteria, across different imaging modalities. Archit et al. \cite{archit2023segment} presented segment anything for microscopy as a tool for interactive and automatic segmentation and tracking of objects in multi-dimensional microscopy data. They extended SAM by training specialized models for microscopy data that significantly improve segmentation quality for a wide range of imaging conditions.

In addition to image segmentation, SAM was also utilized to generate pixel-level annotations to train a segmentation model for pathology images. Li et al. \cite{li2023leverage} investigated the feasibility of bypassing pixel-level delineation through the utilization of SAM applied to weak box annotations in a zero-shot learning framework. Specifically, SAM's capability was leveraged to generate pixel-level annotations from mere box annotations and employed these SAM-derived labels to train a segmentation model. Results demonstrated that the proposed SAM-assisted model significantly reduces the labeling workload for non-expert annotators by relying solely on weak box annotations. A summary of mentioned FM used in pathology image segmentation is presented in Table~\ref{tab1}.A.

To conclude, SAM delivers satisfactory performance on histopathological images, particularly with objects that are sharply defined, and significantly facilitates the annotation process in segmentation tasks where dedicated deep learning models are either unavailable or inaccessible. However, SAM's application in annotating histopathological images encounters several challenges. Firstly, SAM faces difficulties with objects that are interconnected or have indistinct borders (e.g., vascular walls), are prone to prompt ambiguity (e.g., distinguishing between an entire vessel and its lumen), or blend into the background due to low contrast (e.g., sparse tumor cells within an inflammatory backdrop). These limitations can be attributed not only to the absence of microscopic images in the training set but also to the intrinsic characteristics of histopathological images compared to conventional images: (1) the color palette is often limited and similar across different structures, and (2) histopathological tissues are essentially presented on a single plane, contrasting with the three-dimensional perspectives captured in real-world photography, which naturally enhances object delineation. Additionally, generating accurate masks for non-object elements, such as the stroma or interstitial spaces, remains challenging even with extensive input. Technical artifacts such as edge clarification in biopsy samples, and tearing artifacts further impair segmentation performance. Secondly, SAM's overall efficacy falls short of the benchmarks set by SOTA models specifically designed for tasks like nuclei segmentation, particularly in semantic segmentation tasks.

\subsubsection{Pathology Image Classification}\label{PIC1}
Pathology image classification leverages computational methods to categorize and diagnose diseases from medical images acquired during pathology examinations. This process plays a vital role in medical diagnostics, as the precise and prompt classification of diseases can greatly influence the outcomes of patient treatments. The images used in pathology, often derived from biopsies, are intricate, featuring detailed cellular and tissue structures that signify a range of health conditions, such as cancers, inflammatory diseases, and infections. Table~\ref{tab1}.B presents a summary of recent FM used in pathology image classification. The subsequent paragraph provides detailed explanations of these models.

\begin{table}
\caption{ Review of FM for pathology visually prompted (i.e. Image segmentation (A) and image classification (B)) and textually prompted (Image classification (C)) models.}
    \centering
    \label{tab1}
    \resizebox{\textwidth}{!}{%
    \begin{tabular}{@{}p{2.5cm}p{4cm}p{6cm}p{6cm}@{}}
        \toprule
        \textbf{Model} & \textbf{Training Data} & \textbf{Key Features} & \textbf{Outcomes / Performance} \\
 \midrule
        \multicolumn{4}{@{}c@{}}{\textbf{(A) Visually Prompted Image Segmentation}} \\
        \midrule
        SAM \cite{kirillov2023segment}   & \cite{deng2023segment}: Cell nuclei segmentation on WSI & A positive point prompt, 20 point prompts, and comprehensive points or bounding boxes & Exceptional performance in segmenting large, connected objects \\
        \addlinespace
        SAM & \cite{cui2023allin}: Generate pixel-level from points \& bounding boxes. & Label-efficient finetuning of SAM, with no requirement for annotation prompts during inference & Minimizes annotation efforts without compromising on segmentation accuracy \\
        \addlinespace
        SAM  & \cite{zhang2023sampath}: SAM-Path for semantic segmentation of pathology images & Extend SAM by incorporating trainable class prompts, augmented further with a pathology-specific encoder & Improves SAM's capability for performing pathology semantic segmentation \\
        \addlinespace
        SAM & \cite{israel2023foundation}: Cell-SAM for cell segmentation & Extend SAM by introducing CellFinder as a novel prompt engineering technique that cues SAM to produce segmentation & Achieves SOTA performance in segmenting images of mammalian cells \\
        \addlinespace
        SAM & \cite{archit2023segment}: Segmenting objects in multi-dimensional microscopy data & Extend SAM by training specialized models for microscopy data & Significantly improve segmentation quality for a wide range of imaging conditions \\
        \addlinespace
        SAM & \cite{li2023leverage}: Cell segmentation in digital pathology & Use SAM’s ability to produce pixel-level annotations from box annotations to train a segmentation model & Diminish the labeling efforts for lay annotators by only requiring weak box annotations\\
        \midrule
       \multicolumn{4}{@{}c@{}}{\textbf{(B) Visually Prompted Image Classification}} \\
        \midrule
        HIPT \cite{chen2022scaling} & 100M patches from 11,000 WSIs (TCGA) & Hierarchical Image Pyramid Transformer, student-teacher knowledge distillation & Outperforms SOTA for cancer subt6yping \& survival prediction \\
        \addlinespace
        CTransPath \cite{wang2022transformer} & 15M patches from 30,000 WSIs (TCGA \& PAIP) & Swin Transformer with a convolutional backbone & Potential to be a universal model for various histopathological image applications \\
        \addlinespace
        ResNet \cite{ciga2022self} & 25K WSIs, 39K patches (TCGA \& others) & Trained on extensive patch dataset & Combining multiple multi-organ datasets with various types of staining and resolution improves the learned features quality \\
        \addlinespace
        REMEDIS \cite{azizi2023robust} & 29,000 WSIs (TCGA) & SimCLR framework & Improves diagnostic accuracies when compared to supervised baseline models \\
        \addlinespace
        DINOv2 \cite{oquab2023dinov2} & Histopathology data & Vision transformer, \cite{vorontsov2023virchow}: a large model, trained on extensive TCGA dataset, \cite{roth2024low}: Benchmarking against CTransPath and RetCCL as feature extractors & Surpasses SOTA in pan-cancer detection \& subtyping \cite{vorontsov2023virchow}, comparable or better performance with less training \cite{roth2024low} \\
        \addlinespace
        UNI \cite{chen2023general}, ViT-Base \cite{filiot2023scaling} & 100K proprietary slides \cite{chen2023general}, 43M patches from 6,000 WSIs (TCGA)\cite{filiot2023scaling} & \cite{chen2023general}:ViT-large model, 100 million tissue patches, 20 major organ types, \cite{filiot2023scaling}: iBOT framework, surpasses HIPT and CTransPath in TCGA tasks & Generalizable across 33 pathology tasks \cite{chen2023general}, superior performance in TCGA evaluation tasks \cite{filiot2023scaling} \\
        \addlinespace
        Largest Academic Model \cite{campanella2023computational} & 3 billion patches from over 400,000 slides & MAE and DINO pre-training, comparison on pathology data vs. natural images & Superior downstream performance with DINO \\
        \addlinespace
        Rudolfv \cite{dippel2024rudolfv} & Less extensive dataset than competitors & Integrating pathological domain knowledge & Superior performance with fewer parameters \\
        \addlinespace
        Prov-GigaPath \cite{xu2024whole} &  1.3 billion pathology image tiles in 171,189 whole slides from Providence & Vision transformer with dilated self-attention & SOTA performance on various pathology tasks, demonstrating the importance of real-world data and whole-slide modeling \\
        \midrule
        \multicolumn{4}{@{}c@{}}{\textbf{(C) Textually Prompted Image Classification}} \\
        \midrule
        TraP-VQA \cite{naseem2022vision} & PathVQA \cite{he2020pathvqa} & A vision-language transformer for processing pathology images & Outperformed the SOTA comparative methods with public PathVQA dataset \\
        \addlinespace
        Huang et al. \cite{huang2023leveraging} & Image-text paired pathology data sourced from public platforms, including Twitter & A contrastive language-image pre-training model & Promising zero-shot capabilities in classifying new pathological images \\
        \addlinespace
        PathAsst \cite{sun2023pathasst} & ChatGPT/GPT-4 to produce over 180,000 samples & Vicuna-13B language model in conjunction with the CLIP vision encoder & Underscore the capability of leveraging AI-powered generative foundational models to enhance pathology diagnoses \\
        \addlinespace
        MI-Zero \cite{lu2023visual} & 550,000 pathology reports along with other available in-domain text corpora & Reimagines zero-shot transfer within the context of multiple instance learning & Potential usefulness for semi-supervised learning workflows in histopathology \\
        \addlinespace
    \end{tabular}
    }
    \end{table}

        \begin{table}
        \centering
    \label{tab2}
    \resizebox{\textwidth}{!}{%
    \begin{tabular}{@{}p{2.5cm}p{4cm}p{6cm}p{6cm}@{}}
        CONCH \cite{lu2023towards} & 1.17 million image-caption pairs derived from educational materials and PubMed articles & ViT backbone and iBOT self-supervised framework, integrates a vision-language joint embedding space & Potential to directly facilitate machine learning-based workflows requiring minimal or no further supervised fine-tuning \\
        \addlinespace
        CITE \cite{zhang2023text} & PatchGastric stomach tumor pathological image dataset & Language models pre-trained on a wide array of biomedical texts to enhance foundation models for better understanding of pathological images & Achieves leading performance compared with various baselines especially when training data is scarce \\
        \bottomrule
    \end{tabular}
    }
    \end{table}

In recent years, numerous self-supervised techniques for computational pathology image classification have been proposed. For example, Hierarchical Image Pyramid Transformer (HIPT) \cite{chen2022scaling} is a Vision Transformer (ViT) with less than 10 million parameters, trained on approximately 100 million patches extracted from 11,000 WSIs from The Cancer Genome Atlas (TCGA) \cite{weinstein2013cancer}. HIPT utilized student-teacher knowledge distillation \cite{caron2021emerging} at two successive representation levels: initially at the local image patch level and subsequently at the regional image patch level, which is derived from the learned representations of multiple local patches. CTransPath, a Swin Transformer equipped with a convolutional backbone, featuring 28 million parameters was proposed by \cite{wang2022transformer}. It was trained on 15 million patches extracted from 30,000 WSIs sourced from both The TCGA and the Pathology AI Platform (PAIP) \cite{kim2021paip}. Similarly, Ciga et al. \cite{ciga2022self} trained a 45 million parameter ResNet on 25 thousands WSIs with the addition of 39 thousands patches, all collected from TCGA and 56 other small datasets. Recently, Filiot et al. \cite{filiot2023scaling} utilized a ViT-Base architecture, which has 86 million parameters, and employed the Image BERT pre-training with Online Tokenizer (iBOT) framework \cite{zhou2021ibot} for its pre-training. This model was pre-trained on 43 million patches derived from 6,000 WSIs from the TCGA and surpassed both HIPT and CTransPath in performance across various TCGA evaluation tasks. In a similar endeavor, Azizi et al. \cite{azizi2023robust} developed a model with 60 million parameters using the SimCLR framework \cite{chen2020simple} and trained it on 29,000 WSIs, covering nearly the entire TCGA dataset.

The mentioned studies highlight models that possess up to 86 million parameters and incorporate a teacher distillation objective for training on the extensive TCGA dataset, which includes over 30,000 WSIs. In contrast, Virchow \cite{vorontsov2023virchow} distinguishes itself by its significantly larger scale and more extensive training data. It boasts 632 million parameters, marking a 69-fold increase in size compared to the largest models mentioned in the previous studies. Virchow employs the vision transformer and is trained using DINOv2 \cite{oquab2023dinov2} self-supervised algorithm which is based on a student-teacher paradigm. When evaluated on downstream tasks, such as tile-level pan-cancer detection and subtyping, as well as slide-level biomarker prediction, Virchow surpasses SOTA systems. UNI \cite{chen2023general}, a ViT-large model trained on 100 thousand proprietary slides (Mass-100k:  a large and diverse pretraining dataset containing over 100 million tissue patches from 100,426 WSIs across 20 major organ types including normal tissue, cancerous tissue, and other pathologies) is another foundation model for pathology image classification. They assessed the downstream performance on 33 tasks including tile-level tasks for classification, segmentation, and retrieval and slide-level classification tasks and showed generalizability of the model in anatomic pathology. Roth et al. \cite{roth2024low} benchmarked the most popular pathology vision FM like DINOv2 ViT-S, DINOv2 ViT-S finetuned, CTransPath and RetCCL \cite{wang2023retccl} as feature extractors for histopathology data. The models were evaluated in two settings: slide-level classification and patch-level classification. Results showed that finetuning a DINOv2 ViT-S yields at least equal performance compared to CTransPath and RetCCL but in a fraction of domain specific training time. Campanella et al. \cite{campanella2023computational} trained the largest academic foundation model on 3 billion image patches from over 400,000 slides. They compared pre-training of visual transformer models using the masked autoencoder (MAE) and DINO algorithms. The results demonstrate that pre-training on pathology data is beneficial for downstream performance compared to pre-training on natural images. Also, the DINO algorithm achieved better generalization performance across all tasks tested. Furthermore, Dippel et al. \cite{dippel2024rudolfv} demonstrated that integrating pathological domain knowledge carefully can significantly enhance pathology foundation model performance, achieving superior results with the best available performing pathology foundation model. This achievement comes despite using considerably fewer slides and a model with fewer parameters than competing models. Although the mentioned studies successfully applied FM for pathology image analysis, Prov-GigaPath \cite{xu2024whole} stands out from them due to its unique combination of a larger and more diverse dataset (1.3 billion 256×256 pathology image tiles in 171,189 whole slides), and its innovative use of vision transformers with dilated self-attention.
\subsection{Textually Prompted Models}\label{TPM}
Textually prompted models are increasingly recognized as foundational in the field of medical imaging, particularly in computational pathology. These models learn representations that capture the semantics and relationships between pathology images and their corresponding textual prompts (shown in Figure~\ref{fig22}.D). By leveraging contrastive learning objectives, they bring similar image-text pairs closer together in the feature space while pushing dissimilar pairs apart. Such models are crucial for tasks related to pathology image classification and retrieval. Architectural explorations have included dual-encoder designs—with separate visual and language encoders—as well as fusion designs that integrate image and text representations using decoder and transformer-based architectures. The potential of these models for pathology image classification is highlighted in numerous studies, which are discussed in the following section and summarized in Table~\ref{tab1}.C.

\subsubsection{Pathology Image Classification}\label{PIC2}
In the context of pathology image classification, TraP-VQA \cite{naseem2022vision} represents the pioneering effort to utilize a vision-language transformer for processing pathology images. This approach was evaluated using the PathVQA dataset \cite{he2020pathvqa} to generate interpretable answers. More recently, Huang et al. \cite{huang2023leveraging} compiled a comprehensive dataset of image-text paired pathology data sourced from public platforms, including Twitter. They employed a contrastive language-image pre-training model to create a foundational framework for both pathology text-to-image and image-to-image retrieval tasks. Their methodology showcased promising zero-shot capabilities in classifying new pathological images. PathAsst \cite{sun2023pathasst} utilizes FM, functioning as a generative AI assistant, to revolutionize predictive analytics in pathology. It employs ChatGPT/GPT-4 to produce over 180,000 samples that follow instructions, thereby activating pathology-specific models and enabling efficient interactions based on input images and user queries. PathAsst is developed using the Vicuna-13B language model in conjunction with the CLIP vision encoder. The outcomes from PathAsst underscore the capability of leveraging AI-powered generative FM to enhance pathology diagnoses and the subsequent treatment processes. Lu et al. \cite{lu2023visual} introduced MI-Zero, an intuitive framework designed to unlock the zero-shot transfer capabilities of contrastively aligned image and text models for gigapixel histopathology whole slide images. This framework allows multiple downstream diagnostic tasks to be performed using pre-trained encoders without the need for additional labeling. MI-Zero reimagines zero-shot transfer within the context of multiple instance learning, addressing the computational challenges associated with processing extremely large images. To pre-train the text encoder, they utilized over 550,000 pathology reports along with other available in-domain text corpora. By harnessing the power of strong pretrained encoders, their top performing model, which was pretrained on more than 33,000 histopathology image-caption pairs, achieves an average median zero-shot accuracy of 70.2\% across three distinct real-world cancer subtyping tasks. Lu et al. \cite{lu2023towards} proposed CONCH, a foundational model framework for pathology that integrates a vision-language joint embedding space. Initially, they trained a ViT on a dataset comprising 16 million tiles from 21,442 proprietary in-house WSIs using the iBOT \cite{zhou2021ibot} self-supervised learning framework. Subsequently, leveraging the ViT backbone, they developed a vision-language model utilizing the CoCa framework \cite{yu2022coca}, trained on 1.17 million image-caption pairs derived from educational materials and PubMed articles. The model's efficacy was evaluated across 13 downstream tasks, including tile and slide classification, cross-modal image-to-text and text-to-image retrieval, coarse WSI segmentation, and image captioning. Another study \cite{zhang2023text} proposed the Connect Image and Text Embeddings (CITE) method to improve pathological image classification. CITE leverages insights from language models pre-trained on a wide array of biomedical texts to enhance FM for better understanding of pathological images. This approach has shown to achieve superior performance on the PatchGastric stomach tumor pathological image dataset, outperforming various baseline methods particularly in scenarios with limited training data. CITE underscores the value of incorporating domain-specific textual knowledge to bolster efficient pathological image classification.

\section{Endoscopy Foundation Models} \label{EFM}
Endoscopic video has become a standard imaging modality and is increasingly being studied for the diagnosis of gastrointestinal diseases. Developing an effective foundational model shows promise in facilitating downstream tasks requiring analysis of endoscopic videos.

\subsection{Visually Prompted Models}\label{EVPM}
Since clinical routines for endoscopy videos typically do not involve text data, a purely image-based foundational model is currently more feasible. 

In response to this need, Wang et al. \cite{wang2023foundation} developed the first foundation model, Endo-FM, which is specifically designed for analyzing endoscopy videos. Endo-FM utilizes a video transformer architecture to capture rich spatial-temporal information and is pre-trained to be robust against diverse spatial-temporal variations. A large-scale endoscopic video dataset, comprising over 33,000 video clips, was constructed for this purpose. Extensive experimental results across three downstream tasks demonstrate Endo-FM's effectiveness, significantly surpassing other SOTA video-based pre-training methods and showcasing its potential for clinical application. Additionally, Cui et al. \cite{beilei2024surgical} demonstrated the effectiveness of vision-based FM for depth estimation in endoscopic videos. They developed a foundation model-based depth estimation method named Surgical-DINO, which employs a Low-rank Adaptation (LoRA) \cite{hu2021lora} of DINOv2 specifically for depth estimation in endoscopic surgery. The LoRA layers, rather than relying on conventional fine-tuning, were designed and integrated into DINO to incorporate surgery-specific domain knowledge. During the training phase, the image encoder of DINO was frozen to leverage its superior visual representation capabilities, while only the LoRA layers and the depth decoder were optimized to assimilate features from the surgical scene. The results indicated that Surgical-DINO significantly surpasses all other SOTA models in tasks related to endoscopic depth estimation.  Furthermore, trends in the development of video FM indicate promising applications for endoscopy, such as video segmentation \cite{cheng2023segment} for identifying lesions in endoscopy footage, and enhancing endoscopy videos by reconstructing masked information \cite{song2022takes} to reveal obscured lesions.

\section{Challenges and Future Work}\label{CFW}
The pathology and endoscopy FM discussed in this review have their respective shortcomings and open challenges. This section aims to provide a comprehensive overview of the common challenges these approaches face, as well as highlight the future directions of FM in pathology and endoscopy analysis along with FUTURE-AI guidelines that guide their deployments.

Despite the potential of FM for disease diagnosis, their application in the medical domain including pathology and endoscopy image analysis, faces several challenges. Firstly, FM are susceptible to “hallucination” \cite{ji2023survey}, where they generate incorrect or misleading information. In the medical domain, such hallucinations can lead to the dissemination of incorrect medical information, resulting in misdiagnoses and consequently, inappropriate treatments. Secondly, FM in vision and language can inherit and amplify “biases” \cite{hoelscher2023detecting} present in the training data. Biases related to race, underrepresented groups, minority cultures, and gender can result in biased predictions or skewed behavior from the models. Addressing these biases is crucial to ensure fairness, inclusivity, and the ethical deployment of these systems. Additionally, patient privacy and ethical considerations present significant hurdles that must be overcome to ensure the ethical and equitable use of FM in medical practice. Moreover, training large-scale vision and language models demands substantial computational resources and large datasets, which can limit their application in real-time inference or on edge devices with limited computing capabilities. The lack of evaluation benchmarks and metrics for FM also poses a challenge, hindering the assessment of their overall capabilities, particularly in the medical domain. Developing domain-specific and FM-specific benchmarks and metrics is essential.

Despite these challenges, the future direction of FM in pathology and endoscopy analysis is likely to include several innovative and transformative approaches. These models are set to significantly enhance diagnostic accuracy, efficiency, and the overall understanding of disease processes. A key future direction is the integration with multimodal data. FM are expected to evolve beyond text and incorporate the integration of multimodal data, combining pathology images, endoscopic video data, genomic information, and clinical notes. This will enable a more comprehensive and nuanced understanding of patient cases, facilitating more accurate diagnoses and personalized treatment plans. FM could also automate the generation of pathology and endoscopy reports, synthesizing findings from images, patient history, and test results into coherent, standardized, and clinically useful reports. This would streamline workflows, reduce human error, and allow pathologists and endoscopists to concentrate on complex cases. In endoscopy, FM could provide real-time analysis and guidance, identifying areas of interest or concern during a procedure, which could assist less experienced endoscopists and potentially reduce the rate of missed lesions or abnormalities. Future FM are likely to feature more sophisticated algorithms for detecting and classifying diseases from pathology slides and endoscopy videos, trained on vast datasets to recognize rare conditions, subtle abnormalities, and early disease stages with high accuracy.

\begin{figure}[t]\centering
\includegraphics[width=1\textwidth]{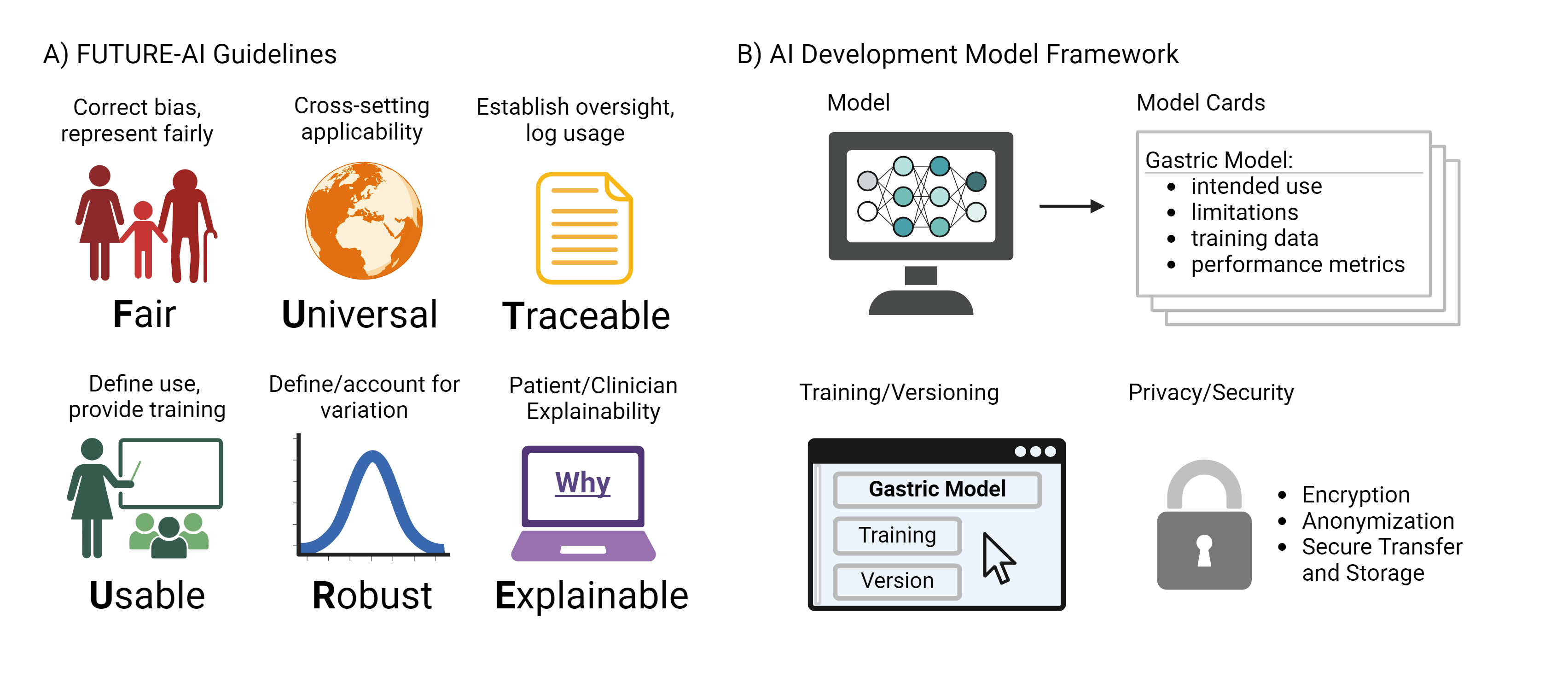}
\caption{A. The FUTURE-AI guidelines. B. Proposed Model Cards Framework.
\label{fig3}}
\end{figure}

Despite their major advances, the deployment and adoption of FM like other medical AI tools remain limited in real-world clinical practice. To increase adoption in the real world, it is essential that medical AI tools are accepted by patients, clinicians, health organizations and authorities. However, there is a lack of widely accepted guidelines on how medical AI tools should be designed, developed, evaluated and deployed to be trustworthy, ethically sound and legally compliant. To address this challenge, the FUTURE-AI guidelines \cite{lekadir2023future} were proposed that aim to guide the development and deployment of AI tools in healthcare that are ethical, legally compliant, technically robust, and clinically safe. It consists of six guiding principles for trustworthy AI including Fairness, Universality, Traceability, Usability, Robustness, and Explainability as shown in Figure~\ref{fig3}.A. This initiative is crucial, especially in domains like gastric cancer detection where AI's potential for early detection and improved patient outcomes is significant. The specificity of vision FM in detecting such conditions necessitates adherence to guidelines ensuring the models' ethical use, fairness, and transparency. By adopting FUTURE-AI's guidelines, researchers and developers can mitigate risks like biases and errors in AI models, ensuring these tools are trustworthy and can be seamlessly integrated into clinical practice. The structured approach provided by FUTURE-AI facilitates the creation of AI tools that are ready for real-world deployment, encouraging their acceptance among patients, clinicians, and health authorities. Additionally, Figure~\ref{fig3}.B shows our proposed AI development framework which provides a structured outline for documenting AI models for the transparency and reliability of gastric cancer detection tools. The Model Cards section details the model's purpose, its limitations, the nature of the training data, and how the model's performance is measured. Training and Versioning are recorded, tracing the evolution of the model through updates and refinements. Privacy and Security considerations are paramount, detailing the protective measures such as encryption and anonymisation to ensure patient data confidentiality during model training and deployment. This multi-dimensional approach to documentation is essential for the end-users and developers to understand the model's capabilities, limitations, and to ensure its responsible use in the future healthcare settings.

\section{Conclusion}
In this survey, we have conducted a comprehensive review of the recent advancements in FM for pathology and endoscopy imaging. Our survey begins with an introductory section, followed by a discussion on the principles of vision FM, including architecture types, training objectives (i.e., contrastive and generative), and large-scale training. Section \ref{PFM} delves into pathology FM, which are classified into visually (section \ref{VPM}) and textually (section \ref{TPM}) prompted models. Visually prompted models are applied to pathology image segmentation and classification, whereas textually prompted models are utilized solely for pathology image classification. Section \ref{EFM} describes recent works on endoscopy FM, which are exclusively visually prompted models. In conclusion, our survey not only reviews recent developments but also lays the groundwork for future research in FM. We propose several directions for future investigations (section \ref{CFW}), offering a roadmap for researchers aiming to excel in the field of FM for pathology and endoscopy imaging.


\acknowledgments{This research is part of AIDA project, funded by the European Union (grant number 101095359) and UK Research and Innovation (grant number 10058099).}




\begin{adjustwidth}{-\extralength}{0cm}

\reftitle{References}

\end{adjustwidth}

\begin{thebibliography}{999}
\bibitem{yoon2015diagnosis}
Yoon, H.; Kim, N. Diagnosis and management of high risk group for gastric cancer. {\em Gut And Liver} \textbf{2015}, \textit{9}, 5.

\bibitem{pimentel2019management}
Pimentel-Nunes, P.; Libânio, D.; Marcos-Pinto, R.; Areia, M.; Leja, M.; Esposito, G.; Garrido, M.; Kikuste, I.; Megraud, F.; Matysiak-Budnik, T; \& Others. Management of epithelial precancerous conditions and lesions in the stomach (maps II): European Society of gastrointestinal endoscopy (ESGE), European Helicobacter and microbiota Study Group (EHMSG), European Society of pathology (ESP), and Sociedade Portuguesa de Endoscopia Digestiva (SPED) guideline update 2019. {\em Endoscopy} \textbf{2019},\textit{51}, 365-388.

\bibitem{matysiak2020recent}
Matysiak-Budnik, T.; Camargo, M.; Piazuelo, M.; Leja, M. Recent guidelines on the management of patients with gastric atrophy: common points and controversies. {\em Digestive Diseases And Sciences} \textbf{2020}, \textit{65}, 1899-1903.

\bibitem{chen2020simple}
Chen, T.; Kornblith, S.; Norouzi, M; \& Hinton, G. A simple framework for contrastive learning of visual representations. {\em International Conference On Machine Learning}. \textbf{2020}, 1597-1607.

\bibitem{radford2021learning}
Radford, A.; Kim, J.; Hallacy, C.; Ramesh, A.; Goh, G.; Agarwal, S.; Sastry, G.; Askell, A.; Mishkin, P.; Clark, J; \& Others. Learning transferable visual models from natural language supervision. {\em International Conference On Machine Learning}. \textbf{2021}, 8748-8763.

\bibitem{jia2021scaling}
Jia, C.; Yang, Y.; Xia, Y.; Chen, Y.-T.; Parekh, Z.; Pham, H.; Le, Q.; Sung, Y.-H.; Li, Z.; and Duerig, T. Scaling up visual and vision-language representation learning with noisy text supervision. In \textit{International Conference on Machine Learning}. \textbf{2021}, 4904–4916.

\bibitem{li2021align}
Li, J.; Selvaraju, R.; Gotmare, A.; Joty, S.; Xiong, C.; and Hoi, S.C.H. Align before fuse: Vision and language representation learning with momentum distillation. \textit{Advances in Neural Information Processing Systems}, 34, \textbf{2021}, 9694–9705.

\bibitem{yao2021filip}
Yao, L.; Huang, R.; Hou, L.; Lu, G.; Niu, M.; Xu, H.; Liang, X.; Li, Z.; Jiang, X.; and Xu, C. Filip: fine-grained interactive language-image pre-training. \textit{arXiv preprint arXiv:2111.07783}, \textbf{2021}.

\bibitem{wang2022cris}
Wang, Z.; Lu, Y.; Li, Q.; Tao, X.; Guo, Y.; Gong, M.; and Liu, T. Cris: Clip-driven referring image segmentation. In \textit{Proceedings of the IEEE/CVF Conference on Computer Vision and Pattern Recognition}, \textbf{2022}, 11686–11695.

\bibitem{li2022grounded}
Li, L.H.; Zhang, P.; Zhang, H.; Yang, J.; Li, C.; Zhong, Y.; Wang, L.; Yuan, L.; Zhang, L.; Hwang, J.-N.; et al. Grounded language-image pre-training. In \textit{Proceedings of the IEEE/CVF Conference on Computer Vision and Pattern Recognition}, \textbf{2022}, 10965–10975.

\bibitem{xu2022groupvit}
Xu, J.; De Mello, S.; Liu, S.; Byeon, W.; Breuel, T.; Kautz, J.; and Wang, X. Groupvit: Semantic segmentation emerges from text supervision. \textit{Computer Vision And Pattern Recognition}, \textbf{2022}.

\bibitem{yang2022unified}
Yang, J.; Li, C.; Zhang, P.; Xiao, B.; Liu, C.; Yuan, L.; and Gao, J. Unified contrastive learning in image-text-label space. In \textit{Proceedings of the IEEE/CVF Conference on Computer Vision and Pattern Recognition}, \textbf{2022}, 19163–19173.

\bibitem{zhang2022glipv2}
Zhang, H.; Zhang, P.; Hu, X.; Chen, Y.-C.; Li, L.; Dai, X.; Wang, L.; Yuan, L.; Hwang, J.-N.; and Gao, J. Glipv2: Unifying localization and vision-language understanding. \textit{Advances in Neural Information Processing Systems}, 35, \textbf{2022}, 36067–36080.

\bibitem{bao2021beit}
Bao, H.; Dong, L.; and Wei, F. Beit: Bert pre-training of image transformers. \textit{arXiv preprint arXiv:2106.08254}, \textbf{2021}.

\bibitem{liu2019roberta}
Liu, Y.; Ott, M.; Goyal, N.; Du, J.; Joshi, M.; Chen, D.; Levy, O.; Lewis, M.; Zettlemoyer, L.; and Stoyanov, V. Roberta: A robustly optimized bert pretraining approach. \textit{arXiv preprint arXiv:1907.11692}, \textbf{2019}.

\bibitem{zhang2023toward}
Zhang, X.; Zeng, Y.; Zhang, J.; and Li, H. Toward building general foundation models for language, vision, and vision-language understanding tasks. \textit{arXiv preprint arXiv:2301.05065}, \textbf{2023}.

\bibitem{singh2022flava}
Singh, A.; Hu, R.; Goswami, V.; Couairon, G.; Galuba, W.; Rohrbach, M.; and Kiela, D. Flava: A foundational language and vision alignment model. In \textit{Proceedings of the IEEE/CVF Conference on Computer Vision and Pattern Recognition}, \textbf{2022}, 15638–15650.


\bibitem{hao2022language}
Hao, Y.; Song, H.; Dong, L.; Huang, S.; Chi, Z.; Wang, W.; Ma, S.; and Wei, F. Language models are general-purpose interfaces. \textit{arXiv preprint arXiv:2206.06336}, \textbf{2022}.

\bibitem{li2023blip2}
Li, J.; Li, D.; Savarese, S.; and Hoi, S. Blip-2: Bootstrapping language-image pre-training with frozen image encoders and large language models. \textit{arXiv preprint arXiv:2301.12597}, \textbf{2023}.

\bibitem{tschannen2023image}
Tschannen, M.; Kumar, M.; Steiner, A.; Zhai, X.; Houlsby, N.; and Beyer, L. Image captioners are scalable vision learners too. \textit{arXiv preprint arXiv:2306.07915}, \textbf{2023}.

\bibitem{chen2020uniter}
Chen, Y.-C.; Li, L.; Yu, L.; El Kholy, A.; Ahmed, F.; Gan, Z.; Cheng, Y.; and Liu, J. Uniter: Universal image-text representation learning. In \textit{Computer Vision–ECCV 2020: 16th European Conference, Part XXX}, \textbf{2020}, 104–120.

\bibitem{tsimpoukelli2021multimodal}
Tsimpoukelli, M.; Menick, J.L.; Cabi, S.; Eslami, S.M.; Vinyals, O.; and Hill, F. Multimodal few-shot learning with frozen language models. \textit{Advances in Neural Information Processing Systems}, 34, \textbf{2021}, 200–212.

\bibitem{xu2022unifying}
Xu, H.; Zhang, J.; Cai, J.; Rezatofighi, H.; Yu, F.; Tao, D.; and Geiger, A. Unifying flow, stereo and depth estimation. \textit{arXiv preprint arXiv:2211.05783}, \textbf{2022}.

\bibitem{kirillov2023segment}
Kirillov, A.; Mintun, E.; Ravi, N.; Mao, H.; Rolland, C.; Gustafson, L.; Xiao, T.; et al. Segment anything. In \textit{Proceedings of the IEEE/CVF International Conference on Computer Vision}, \textbf{2023}, 4015–4026.

\bibitem{deng2023segment}
Deng, R.; Cui, C.; Liu, Q.; Yao, T.; Remedios, L.W.; Bao, S.; Landman, B.A.; Wheless, L.E.; Coburn, L.A.; Wilson, K.T.; et al. Segment anything model (sam) for digital pathology: Assess zero-shot segmentation on whole slide imaging. \textit{arXiv preprint arXiv:2304.04155}, \textbf{2023}.

\bibitem{cui2023allin}
Cui, C.; Deng, R.; Liu, Q.; Yao, T.; Bao, S.; Remedios, L.W.; Tang, Y.; and Huo, Y. All-in-sam: from weak annotation to pixel-wise nuclei segmentation with prompt-based finetuning. \textit{arXiv preprint arXiv:2307.00290}, \textbf{2023}.

\bibitem{zhang2023sampath}
Zhang, J.; Ma, K.; Kapse, S.; Saltz, J.; Vakalopoulou, M.; Prasanna, P.; and Samaras, D. Sam-path: A segment anything model for semantic segmentation in digital pathology. \textit{arXiv preprint arXiv:2307.09570}, \textbf{2023}.

\bibitem{israel2023foundation}
Israel, U.; Marks, M.; Dilip, R.; Li, Q.; Schwartz, M.S.; Pradhan, E.; Pao, E.; et al. A foundation model for cell segmentation. \textit{bioRxiv}, \textbf{2023}, 2023-11.

\bibitem{archit2023segment}
Archit, A.; Nair, S.; Khalid, N.; Hilt, P.; Rajashekar, V.; Freitag, M.; Gupta, S.; Dengel, A.; Ahmed, S.; and Pape, C. Segment anything for microscopy. \textit{bioRxiv}, \textbf{2023}, 2023-08.

\bibitem{li2023leverage}
Li, X.; Deng, R.; Tang, Y.; Bao, S.; Yang, H. and Huo, Y. Leverage Weakly Annotation to Pixel-wise Annotation via Zero-shot Segment Anything Model for Molecular-empowered Learning. \textit{arXiv preprint arXiv:2308.05785}, \textbf{2023}.

\bibitem{chen2022scaling}
Chen, R.J.; Chen, C.; Li, Y.; Chen, T.Y.; Trister, A.D.; Krishnan, R.G.; Mahmood, F. Scaling vision transformers to gigapixel images via hierarchical self-supervised learning. In \textit{Proceedings of the IEEE/CVF Conference on Computer Vision and Pattern Recognition}, \textbf{2022}, 16144--16155.

\bibitem{weinstein2013cancer}
Weinstein, J.N.; Collisson, E.A.; Mills, G.B.; Shaw, K.R.; Ozenberger, B.A.; Ellrott, K.; Shmulevich, I.; Sander, C.; Stuart, J.M. The cancer genome atlas pan-cancer analysis project. \textit{Nature Genetics}, \textbf{2013}, 45(10): 1113--1120.

\bibitem{caron2021emerging}
Caron, M.; Touvron, H.; Misra, I.; Jégou, H.; Mairal, J.; Bojanowski, P.; Joulin, A. Emerging properties in self-supervised vision transformers. In \textit{Proceedings of the IEEE/CVF International Conference on Computer Vision}, \textbf{2021}, 9650--9660.

\bibitem{wang2022transformer}
Wang, X.; Yang, S.; Zhang, J.; Wang, M.; Zhang, J.; Yang, W.; Huang, J.; Han, X. Transformer-based unsupervised contrastive learning for histopathological image classification. \textit{Medical Image Analysis}, \textbf{2022}, 81: 102559.

\bibitem{kim2021paip}
Kim, Y.J.; Jang, H.; Lee, K.; Park, S.; Min, S.-G.; Hong, C.; Park, J.H.; Lee, K.; Kim, J.; Hong, W.; et al. Paip 2019: Liver cancer segmentation challenge. \textit{Medical Image Analysis}, \textbf{2021}, 67:101854.

\bibitem{ciga2022self}
Ciga, O.; Xu, T.; Martel, A.L. Self supervised contrastive learning for digital histopathology. \textit{Machine Learning with Applications}, \textbf{2022}, 7:100198.

\bibitem{filiot2023scaling}
Filiot, A.; Ghermi, R.; Olivier, A.; Jacob, P.; Fidon, L.; Mac Kain, A.; Saillard, C.; Schiratti, J.-B. Scaling self-supervised learning for histopathology with masked image modeling. \textit{medRxiv}, \textbf{2023}, 2023--07.

\bibitem{zhou2021ibot}
Zhou, J.; Wei, C.; Wang, H.; Shen, W.; Xie, C.; Yuille, A.; Kong, T. ibot: Image bert pre-training with online tokenizer. \textit{arXiv preprint arXiv:2111.07832}, \textbf{2021}.

\bibitem{azizi2023robust}
Azizi, S.; Culp, L.; Freyberg, J.; Mustafa, B.; Baur, S.; Kornblith, S.; Chen, T.; Tomasev, N.; Mitrović, J.; Strachan, P.; et al. Robust and data-efficient generalization of self-supervised machine learning for diagnostic imaging. \textit{Nature Biomedical Engineering}, \textbf{2023}, 1--24.

\bibitem{oquab2023dinov2}
Oquab, M.; Darcet, T.; Moutakanni, T.; Vo, H.; Szafraniec, M.; Khalidov, V.; Fernandez, P.; Haziza, D.; Massa, F.; El-Nouby, A.; et al. Dinov2: Learning robust visual features without supervision. \textit{arXiv preprint arXiv:2304.07193}, \textbf{2023}.

\bibitem{vorontsov2023virchow}
Vorontsov, E.; Bozkurt, A.; et al. Virchow: A million-slide digital pathology foundation model. \textit{arXiv preprint arXiv:2309.07778}, \textbf{2023}.

\bibitem{chen2023general}
Chen, R.J.; Ding, T.; Lu, M.Y.; Williamson, D.F.K.; Jaume, G.; Chen, B.; Zhang, A.; et al. A general-purpose self-supervised model for computational pathology. \textit{arXiv preprint arXiv:2308.15474}, \textbf{2023}.

\bibitem{roth2024low}
Roth, B.; Koch, V.; Wagner, S.J.; Schnabel, J.A.; Marr, C.; Peng, T. Low-resource finetuning of foundation models beats state-of-the-art in histopathology. \textit{arXiv preprint arXiv:2401.04720}, \textbf{2024}.

\bibitem{wang2023retccl}
Wang, X.; Du, Y.; et al. Retccl: clustering-guided contrastive learning for whole-slide image retrieval. \textit{Medical Image Analysis}, \textbf{2023}, vol. 83, 102645.

\bibitem{campanella2023computational}
Campanella, G.; Kwan, R.; Fluder, E.; Zeng, J.; Stock, A.; Veremis, B.; Polydorides, A.D.; Hedvat, C.; Schoenfeld, A.; Vanderbilt, C.; et al. Computational pathology at health system scale–self-supervised foundation models from three billion images. \textit{arXiv preprint arXiv:2310.07033}, \textbf{2023}.

\bibitem{dippel2024rudolfv}
Dippel, J.; Feulner, B.; Winterhoff, T.; Schallenberg, S.; Dernbach, G.; Kunft, A.; Tietz, S.; Jurmeister, P.; Horst, D.; Ruff, L.; Müller, K.R. RudolfV: A Foundation Model by Pathologists for Pathologists. \textit{arXiv preprint arXiv:2401.04079}, \textbf{2024}.

\bibitem{xu2024whole}
Xu, H.; Usuyama, N.; Bagga, J.; Zhang, S.; Rao, R.; Naumann, T.; Wong, C.; et al. A whole-slide foundation model for digital pathology from real-world data. \textit{Nature}, \textbf{2024}, 1--8.

\bibitem{naseem2022vision}
Naseem, U.; Khushi, M.; Kim, J. Vision-language transformer for interpretable pathology visual question answering. \textit{IEEE Journal of Biomedical and Health Informatics}, \textbf{2022}.

\bibitem{he2020pathvqa}
He, X.; Zhang, Y.; Mou, L.; Xing, E.; Xie, P. Pathvqa: 30000+ questions for medical visual question answering. \textit{arXiv preprint arXiv:2003.10286}, \textbf{2020}.

\bibitem{huang2023leveraging}
Huang, Z.; Bianchi, F.; Yuksekgonul, M.; Montine, T.; Zou, J. Leveraging medical twitter to build a visual–language foundation model for pathology ai. \textit{bioRxiv}, \textbf{2023}, 2023--03.

\bibitem{sun2023pathasst}
Sun, Y.; Zhu, C.; Zheng, S.; Zhang, K.; Shui, Z.; Yu, X.; Zhao, Y.; Li, H.; Zhang, Y.; Zhao, R.; Lyu, X. Pathasst: Redefining pathology through generative foundation ai assistant for pathology. \textit{arXiv preprint arXiv:2305.15072}, \textbf{2023}.

\bibitem{lu2023visual}
Lu, M.Y.; Chen, B.; Zhang, A.; Williamson, D.F.; Chen, R.J.; Ding, T.; Le, L.P.; Chuang, Y.S.; Mahmood, F. Visual Language Pretrained Multiple Instance Zero-Shot Transfer for Histopathology Images. In \textit{Proceedings of the IEEE/CVF Conference on Computer Vision and Pattern Recognition}, \textbf{2023}.

\bibitem{lu2023towards}
Lu, M.Y.; Chen, B.; Williamson, D.F.; Chen, R.J.; Liang, I.; Ding, T.; Jaume, G.; Odintsov, I.; Zhang, A.; Le, L.P.; Gerber, G. Towards a visual-language foundation model for computational pathology. \textit{arXiv preprint arXiv:2307.12914}, \textbf{2023}.

\bibitem{yu2022coca}
Yu, J.; et al. Coca: Contrastive captioners are image-text foundation models. \textit{arXiv preprint arXiv:2205.01917}, \textbf{2022}.

\bibitem{zhang2023text}
Zhang, Y.; Gao, J.; Zhou, M.; Wang, X.; Qiao, Y.; Zhang, S.; Wang, D. Text-guided foundation model adaptation for pathological image classification. In \textit{International Conference on Medical Image Computing and Computer-Assisted Intervention}, \textbf{2023}, 272--282.

\bibitem{wang2023foundation}
Wang, Z.; Liu, C.; Zhang, S.; Dou, Q. Foundation model for endoscopy video analysis via large-scale self-supervised pre-train. In \textit{International Conference on Medical Image Computing and Computer-Assisted Intervention}, \textbf{2023}, 101--111.

\bibitem{beilei2024surgical}
Cui, B.; Mobarakol, I.; Long, B.; Hongliang, R. Surgical-DINO: Adapter Learning of Foundation Model for Depth Estimation in Endoscopic Surgery. \textit{arXiv preprint arXiv:2401.06013}, \textbf{2024}.

\bibitem{hu2021lora}
Hu, E.J.; Shen, Y.; Wallis, P.; Allen-Zhu, Z.; Li, Y.; Wang, S.; Wang, L.; Chen, W. Lora: Low-rank adaptation of large language models. \textit{arXiv preprint arXiv:2106.09685}, \textbf{2021}.

\bibitem{cheng2023segment}
Cheng, Y.; Li, L.; Xu, Y.; Li, X.; Yang, Z.; Wang, W.; Yang, Y. Segment and track anything. \textit{arXiv preprint arXiv:2305.06558}, \textbf{2023}.

\bibitem{song2022takes}
Song, Y.; Yang, M.; Wu, W.; He, D.; Li, F.; Wang, J. It takes two: Masked appearance-motion modeling for self-supervised video transformer pre-training. \textit{arXiv preprint arXiv:2022}, \textbf{2022}.

\bibitem{ji2023survey}
Ji, Z.; Lee, N.; Frieske, R.; Yu, T.; Su, D.; Xu, Y.; Ishii, E.; Bang, Y.J.; Madotto, A.; Fung, P. Survey of hallucination in natural language generation. \textit{ACM Computing Surveys}, \textbf{2023}, 55(12):1--38.

\bibitem{hoelscher2023detecting}
Hoelscher-Obermaier, J.; Persson, J.; Kran, E.; Konstas, I.; Barez, F. Detecting edit failures in large language models: An improved specificity benchmark. \textit{arXiv preprint arXiv:2305.17553}, \textbf{2023}.

\bibitem{lekadir2023future}
Lekadir, K.; Feragen, A.; Fofanah, AJ.; Frangi, A.; Buyx, A.; Emelie, A.; Lara, A.; et al. FUTURE-AI: International consensus guideline for trustworthy and deployable artificial intelligence in healthcare. \textit{arXiv preprint arXiv:2309.12325}, \textbf{2023}.
\end{thebibliography}
\end{document}